\documentclass{article}

\usepackage[final,main]{neurips_2026}

\usepackage[utf8]{inputenc}
\usepackage[T1]{fontenc}
\usepackage{hyperref}
\usepackage{url}
\usepackage{booktabs}
\usepackage{amsfonts}
\usepackage{amsmath}
\usepackage{nicefrac}
\usepackage{microtype}
\usepackage{graphicx}
\usepackage{xcolor}
\usepackage{algorithm}
\usepackage{algorithmic}
\usepackage{subcaption}
\usepackage{wrapfig}
\usepackage{multirow}
\usepackage{listings}
\usepackage{enumitem}
\usepackage{graphicx}

\lstdefinestyle{python}{
  language=Python,
  basicstyle=\ttfamily\small,
  keywordstyle=\color[rgb]{0.0,0.5,0.0}\bfseries,
  stringstyle=\color[rgb]{0.75,0.0,0.0},
  commentstyle=\color[rgb]{0.5,0.5,0.5}\itshape,
  numberstyle=\tiny\color{gray},
  numbers=left,
  numbersep=6pt,
  xleftmargin=2em,
  framexleftmargin=1.5em,
  breaklines=true,
  showstringspaces=false,
  tabsize=4,
  columns=fullflexible,
  keepspaces=true,
  morekeywords={Tensor, Linear, RMSNorm, list, tuple, self, None},
}

\title{Delta Attention Residuals}

\author{
  Cheng Luo\thanks{Equal contribution. Corresponding authors: \texttt{wdlctc@gmail.com}, \texttt{zefncai@gmail.com}, \texttt{junjie.hu@wisc.edu}.} \\
  Independent Researcher \\
  \texttt{wdlctc@gmail.com} \\
  \And
  Zefan Cai\footnotemark[1] \\
  University of Wisconsin--Madison \\
  \texttt{zefncai@gmail.com} \\
  \And
  Junjie Hu \\
  University of Wisconsin--Madison \\
  \texttt{junjie.hu@wisc.edu} \\
}

\begin{document}

\maketitle

\begin{abstract}
Attention Residuals replace standard additive residual connections with learned softmax attention over previous layer outputs, enabling selective cross-layer routing. However, standard Attention Residuals still attend over cumulative hidden states in previous layers, which are highly redundant. We show that this redundancy leads to \textit{routing collapse} in deeper layers: attention weights become low-contrast and closer to uniform (max weight ${\approx}$0.2), limiting the model's ability to select informative states in previous layers. This raises a key but underexplored design question: \textbf{what layer-wise representations should be routed in Attention Residuals}? To answer this question, we propose \textbf{Delta Attention Residuals}, which attend over \emph{deltas}---the change introduced by each sublayer ($\mathbf{v}_i = \mathbf{h}_{i+1} - \mathbf{h}_i$)---instead of cumulative states. Delta representations are structurally diverse and yield higher-contrast attention distributions (max weight ${\approx}$0.6), enabling more selective and effective routing across layers. This principle applies at both per-sublayer and block granularity. Across all tested scales (220M--7.6B), Delta Attention Residuals consistently outperform both standard residuals and Attention Residuals, with 1.7--8.2\% validation perplexity gains. Delta Attention Residuals also enables converting pretrained checkpoints into Delta Attention Residuals via standard fine-tuning. 
Code is available at \url{https://github.com/wdlctc/delta-attention-residuals-code}.
\end{abstract}

%% ============================================================
\section{Introduction}
%% ============================================================

Residual connections~\citep{he2016deep} are fundamental to training deep transformers~\citep{vaswani2017attention}. The standard update, $\mathbf{h}_l = \mathbf{h}_{l-1} + f_l(\mathbf{h}_{l-1})$, accumulates all preceding sublayer outputs with fixed additive coefficients. While this update provides gradient highways~\citep{veit2016residual}, it has no mechanism to selectively aggregate the preceding layers across depth. Attention Residuals~\citep{kimi2025attention} address this issue by replacing fixed aggregation with learned softmax attention over prior layer outputs ($\mathbf{h}_0, \mathbf{h}_1, \ldots, \mathbf{h}_{1-1}$), allowing selective residuals routed from prior layers to the current layer.%achieving improved scaling laws at up to 16B parameters.

However, as each state $\mathbf{h}_l$ in Attention Residuals is still a running sum of all previous layer outputs $\mathbf{h}_{0:l-1}$, adjacent states $\mathbf{h}_l, \mathbf{h}_{l-1}$ become highly redundant. As depth increases, the pairwise similarity between adjacent states grows, reducing the contrast among routing candidates. Under such low-contrast sources, softmax attention becomes less discriminative, leading to \emph{routing collapse}, where attention weights approach a near-uniform distribution. Empirically, we observe that the maximum routing weight drops to about 0.2 in deep layers (Figure~\ref{fig:design_space}a), indicating that the mechanism loses its ability to meaningfully select among sources and instead averages over them.

% Yet a critical design choice remains underexplored: \textbf{what should the model attend over?} Standard Attention Residuals route over cumulative hidden states $\mathbf{h}_0, h_1, \ldots$, but since each state is a running sum of all previous sublayer outputs, adjacent states are highly redundant. This causes routing to collapse in deep layers: the max softmax weight drops to ${\sim}$0.2 (Figure~\ref{fig:design_space}a), and the routing mechanism cannot meaningfully differentiate between sources.

This raises a critical yet underexplored design question: \textbf{what layer-wise sources should be routed to the current layer?} 
We observe that the \emph{change} each layer introduces is far more informative than the cumulative states. The delta $\mathbf{v}_i = \mathbf{h}_{i+1} - \mathbf{h}_i$ captures what a specific sublayer \emph{contributed}, not where the model \emph{has been}. Adjacent deltas are naturally diverse because delta outputs serve different functions and operate in different subspaces~\citep{elhage2021mathematical}, while cumulative states converge to near-linear relationships across layers~\citep{razzhigaev2024linear}.

We propose \textbf{Delta Attention Residuals}, which route over these deltas instead of cumulative states. The same principle applies at two granularities: per-sublayer deltas (each attention or MLP output individually) and block-level deltas (the accumulated change over a group of layers). Delta routing uses an \emph{additive} formulation $(\mathbf{h} = \tilde{\mathbf{h}} + \sum \alpha_i \mathbf{v}_i)$. This enables sharp, selective cross-layer shortcuts with max softmax weight increased to ${\sim}$0.6 (Figure~\ref{fig:design_space}a) and the additive formulation preserves the residual stream and enables finetuning of pretrained models with zero initialization disruption.

Our contributions are:

\begin{figure}[t]
\centering
\includegraphics[width=\textwidth]{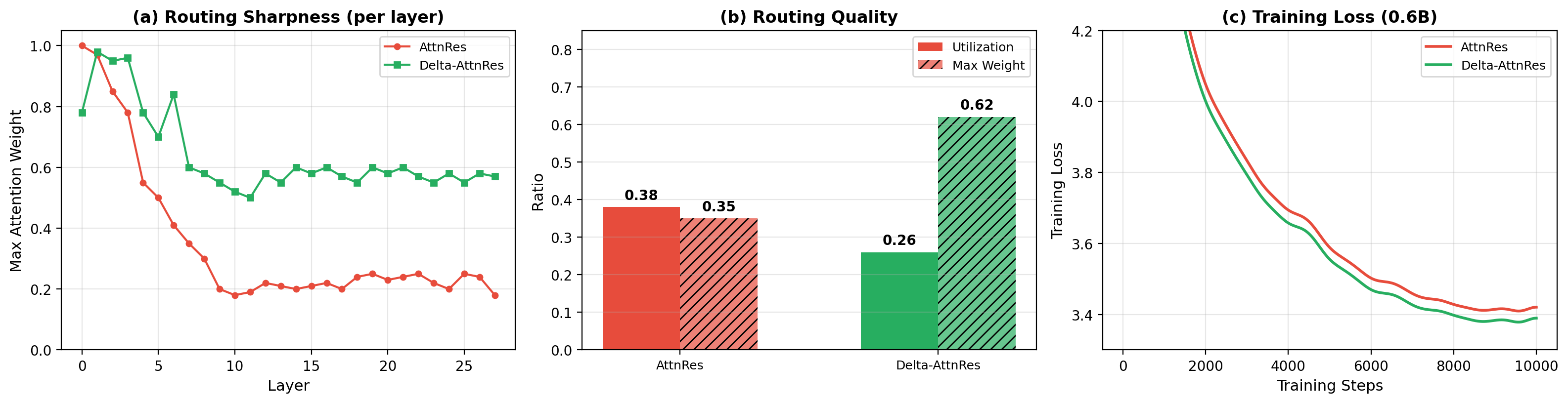}
\vspace{-0.15in}
\caption{\textbf{Source redundancy in cross-layer routing} (Qwen3-0.6B, $L{=}28$). (a)~Per-layer routing sharpness: AttnRes routing degrades to max weight ${\sim}$0.2 in deep layers, while Delta Block maintains sharp routing (${\sim}$0.6). (b)~Routing quality: Delta Block achieves $1.8{\times}$ higher average max weight (0.62 vs.\ 0.35). (c)~Training loss: Delta Block (green) consistently outperforms AttnRes (red).}
\label{fig:design_space}
\vspace{-0.15in}
\end{figure}

\begin{enumerate}[leftmargin=20pt]
    \item We identify the \textbf{routing collapse problem} in Attention Residuals: source redundancy causes routing sharpness to degrade to max weight ${\sim}$0.2 in deep layers, rendering the mechanism near-uniform (\S\ref{sec:what}).
    \item We propose \textbf{Delta Attention Residuals}, which route over deltas ($\mathbf{v}_i = \mathbf{h}_{i+1} - \mathbf{h}_i$) instead of cumulative states, maintaining sharp routing (max weight ${\sim}$0.6) and consistently improving over both baseline and AttnRes from 220M to 7.6B parameters Qwen-based model, (\S\ref{sec:experiments}).
    \item We show that delta routing's \textbf{additive formulation} enables the easy conversion of existing pretrained transformers into Delta Attention Residuals via fine-tuning. Fine-tuned Delta Block outperforms their pretrained transformer checkpoints on 8 downstream benchmarks  (\S\ref{sec:finetune}).
\end{enumerate}

%% ============================================================
\section{Method: Delta Attention Residuals}
\label{sec:method}
%% ============================================================

\subsection{Preliminaries: Attention Residuals}

A Pre-Norm transformer~\citep{vaswani2017attention,xiong2020prenorm} updates the hidden state as $\mathbf{h}_{i+1} = \mathbf{h}_i + \mathbf{v}_i$, where $\mathbf{v}_i := f_i(\mathbf{h}_i)$ is defined as the output of sublayer $i$---here a sublayer can be either an attention or MLP layer. Equivalently, $\mathbf{v}_i = \mathbf{h}_{i+1} - \mathbf{h}_i$ also measures the change introduced at depth $i$. Standard residuals accumulate all sublayer outputs with fixed unit coefficients. Attention Residuals~\citep{kimi2025attention} replace this with a learned weighted sum:
\begin{equation}
    \hat{\mathbf{h}}_l = \sum_{i=0}^{l-1} \alpha_{i \to l} \cdot \mathbf{s}_i, \quad
    \alpha_{i \to l} = \frac{\exp(\mathbf{w}_l^\top \mathrm{RMSNorm}(\mathbf{s}_i))}{\sum_{j} \exp(\mathbf{w}_l^\top \mathrm{RMSNorm}(\mathbf{s}_j))}
    \label{eq:attnres}
\end{equation}
where $\mathbf{s}_i$ are the \emph{source} representations from preceding layers $i\in[0,l)$, $\mathbf{w}_l \in \mathbb{R}^d$ is a learned query (zero-initialized), and $\hat{\mathbf{h}}_l$ feeds into the next sublayer. A critical question arises: \emph{what should $\mathbf{s}_i$ be?}

\subsection{The Source Redundancy Problem}
\label{sec:what}

\begin{figure}[t]
\centering
\includegraphics[width=\textwidth]{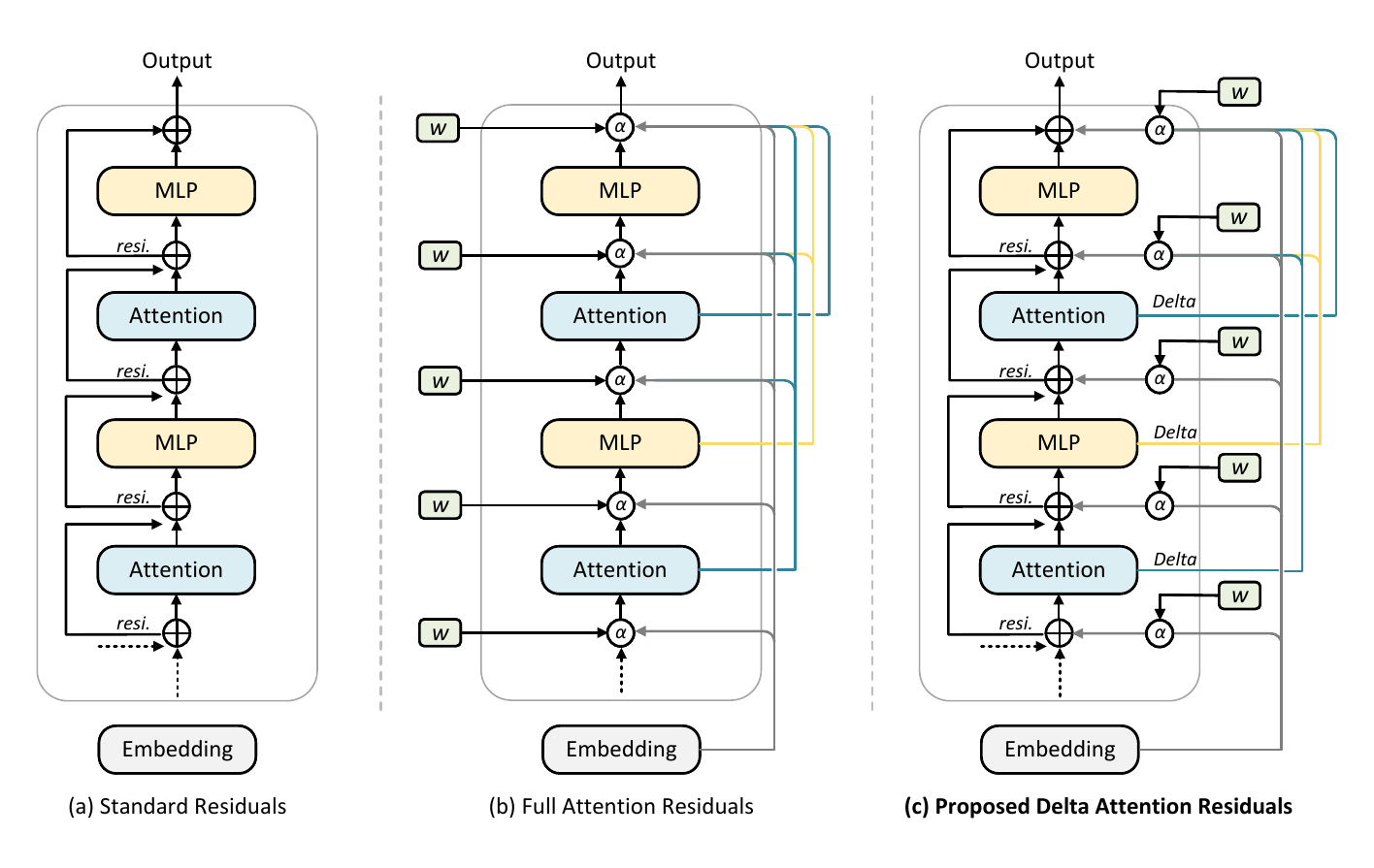}
\vspace{-0.15in}
\caption{\textbf{Architecture comparison.} (a)~Standard Residuals: uniform additive accumulation with fixed unit coefficients. (b)~Attention Residuals~\citep{kimi2025attention}: learned softmax attention (weights $\mathbf{w}$, aggregation $a$) over cumulative hidden states, but source redundancy degrades routing to max weight ${\sim}$0.2 in deep layers. (c)~\textbf{Delta Attention Residuals} (ours): attends over per-sublayer delta outputs, maintaining sharp routing (max weight ${\sim}$0.6) via additive routing. Compatible with MLP layers.}
\vspace{-0.15in}
\label{fig:architecture}
\end{figure}

We consider two source representations for $\mathbf{s}_i$ in Eq.~\ref{eq:attnres}:

\paragraph{Cumulative sources.} $\mathbf{s}_i = \mathbf{h}_i = \mathbf{h}_0 + \sum_{j=1}^{i} \mathbf{v}_j$---the full hidden state after sublayer $i$. Since each $\mathbf{h}_i$ is a running sum, adjacent states share an increasingly large common prefix as depth grows, so the softmax logits $\mathbf{w}^\top \mathrm{RMSNorm}(\mathbf{h}_i) \approx \mathbf{w}^\top \mathrm{RMSNorm}(\mathbf{h}_{i+1})$ and the routing distribution approaches uniform. Empirically, at Qwen3-0.6B scale ($L{=}28$), the max softmax weight drops to ${\sim}$0.2 in deep layers (Figure~\ref{fig:design_space}a), rendering the mechanism near-uniform.

\paragraph{Delta sources.} $\mathbf{s}_i = \mathbf{v}_i$---the per-sublayer output defined in \S\ref{sec:method}. At the finest granularity, each attention output and each MLP output is a separate source (\textbf{Delta AttnRes}, $2L$ sources for $L$ layers). Adjacent deltas are structurally diverse because attention and MLP outputs occupy different subspaces, and outputs from different depths capture different abstraction levels. This naturally coarsens to block-level grouping: a block delta $\Delta_b = \mathbf{h}_{b+1} - \mathbf{h}_b$ aggregates multiple sublayer outputs into one source (\textbf{Delta Block}, $B{+}1$ sources for $B$ blocks), trading granularity for efficiency. See Appendix~\ref{app:block} for a detailed schematic of the block variant.

\subsection{Delta Attention Residuals}

Delta Attention Residuals use \emph{additive routing}: rather than replacing the residual stream with a weighted combination of cumulative states, we \emph{add} selected delta information:
\begin{equation}
    \hat{\mathbf{h}}_l = \tilde{\mathbf{h}}_l + \sum_{i} \alpha_{i \to l} \cdot \mathbf{v}_i
    \label{eq:delta_routing}
\end{equation}
where $\mathbf{v}_i$ are per-sublayer outputs, $\tilde{\mathbf{h}}_l$ is the current residual stream, and $\alpha_{i \to l} = \mathrm{softmax}(\mathbf{w}_l^\top \mathrm{RMSNorm}(\mathbf{v}_i))$. This formulation has three advantages:
\begin{enumerate}[leftmargin=20pt]
    \item \textbf{Residual preservation.} The residual stream $\tilde{\mathbf{h}}_l$ is preserved by default; routing adds information rather than replacing it.
    \item \textbf{No information loss at block boundaries.} In cumulative-state AttnRes, only the states at block boundaries are retained as routing sources---intermediate sublayer contributions within each block are collapsed into a single sum and become individually inaccessible. Delta routing retains every sublayer's contribution as a distinct source, ensuring that no intermediate computation is lost to aggregation.
    \item \textbf{Safe initialization.} At initialization ($\mathbf{w}_l = 0$), all input logits are zero and softmax produces uniform weights, so the routing output $\sum \alpha_i \mathbf{v}_i$ is a bounded perturbation. This makes \texttt{depth\_route} reduce to the identity map on $\tilde{\mathbf{h}}_l$, enabling disruption-free fine-tuning of pretrained models (\S\ref{sec:finetune}).
\end{enumerate}

Figure~\ref{fig:code} shows the complete implementation;  Comparision between  with the original AttnRes can be gound Appendix~\ref{app:comparison}.

\begin{figure}[t]
\begin{lstlisting}[style=python]
def depth_route(sources: list[Tensor], residual: Tensor,
                proj: Linear, norm: RMSNorm) -> Tensor:
    """Softmax attention over depth sources, added to residual."""
    V = torch.stack(sources)                       # [N, B, T, D]
    K = norm(V)
    logits = einsum('d, n b t d -> n b t', proj.weight.squeeze(), K)
    return residual + einsum('n b t, n b t d -> b t d', logits.softmax(0), V)

def forward(self, deltas, hidden_states):
    # --- Attention sublayer ---
    h = depth_route(deltas, hidden_states, self.attn_proj, self.attn_norm)
    attn_out = self.attn(norm(h))
    hidden_states = hidden_states + attn_out

    if self.mode == "Delta Block":
        if not deltas:                             # first layer: store embed
            deltas.append(hidden_states - attn_out)
        deltas.append(hidden_states - deltas[-1])  # store block delta
    elif self.mode == "Delta AttnRes":
        deltas.append(attn_out)                    # store attention delta

    # --- MLP sublayer ---
    h = depth_route(deltas, hidden_states, self.mlp_proj, self.mlp_norm)
    mlp_out = self.mlp(norm(h))
    hidden_states = hidden_states + mlp_out

    if self.mode == "Delta AttnRes":
        deltas.append(mlp_out)                     # store MLP delta

    return deltas, hidden_states
\end{lstlisting}
\caption{\textbf{Delta Attention Residuals pseudocode.} \texttt{depth\_route} computes additive softmax routing over delta sources. \textbf{Delta Block} stores the embedding on first call and appends block deltas (change since last source); \textbf{Delta AttnRes} appends all sublayer outputs directly.}
\label{fig:code}

\vspace{-0.15in}
\end{figure}

%% ============================================================
\section{Experiments}
\label{sec:experiments}
%% ============================================================

\subsection{Experimental Setup}

We train Qwen3-architecture~\citep{qwen2025qwen3} models from scratch on FineWeb-Edu~\citep{penedo2024fineweb}:
\begin{itemize}[leftmargin=20pt]
    \item \textbf{Scales}: 220M ($d{=}768$, $L{=}12$), 533M ($d{=}1024$, $L{=}24$), 1044M ($d{=}1280$, $L{=}36$), plus standard Qwen3-0.6B ($d{=}1024$, $L{=}28$) and Qwen3-8B ($d{=}4096$, $L{=}36$) configurations
    \item \textbf{Training}: AdamW~\citep{loshchilov2019adamw} ($\beta_1{=}0.9$, $\beta_2{=}0.95$, wd${=}0.1$), cosine LR with 500-step warmup, lr $6{\times}10^{-4}$ (220M--1044M) or $3{\times}10^{-4}$ (8B)
    \item \textbf{Budget}: 10K steps, effective batch size 32 (220M--1044M) or 64 (0.6B, 8B), sequence length 1024 (220M--1044M) or 2048 (0.6B, 8B)
    \item \textbf{Hardware}: 8$\times$ NVIDIA H100 80GB, BF16 mixed precision, \texttt{torch.compile}
\end{itemize}

All models are compiled with \texttt{torch.compile} (default mode) before DDP wrapping, and trained with \texttt{use\_cache=False} to avoid unnecessary KV cache allocation during training. Throughput (Tok/s) is measured as total tokens across all GPUs per wall-clock second at steady state. Peak memory (Mem) is the per-device \texttt{torch.cuda.max\_memory\_allocated} during training (batch size 4 per GPU, seq len 1024).

We evaluate five configurations: \textbf{Baseline} (standard residual), \textbf{AttnRes} (cumulative block-level sources with replacement routing, following~\citet{kimi2025attention}), \textbf{Full AttnRes} (AttnRes with $N{=}L$, i.e.\ each layer as its own block), \textbf{Delta AttnRes} (per-sublayer delta sources with additive routing), and \textbf{Delta Block} (delta sources with block-level grouping and additive routing). Since the original Attention Residuals implementation is not publicly available, AttnRes and Full AttnRes are our faithful reimplementation based on the description in~\citet{kimi2025attention} (see Appendix~\ref{app:comparison} for details). All AttnRes variants use zero-initialized queries and identical hyperparameters per scale.

\subsection{From-Scratch Training Results}

\begin{table}[t]
\centering
\caption{Results across three model scales (10K steps, FineWeb-Edu). All methods share identical architecture, data, and hyperparameters per scale; only the depth-mixing mechanism differs. Best per scale in \textbf{bold}. Tok/s measured on 8$\times$H100 (BF16, \texttt{torch.compile}); Mem is peak per-device GPU memory during training (batch size 4 per GPU, seq len 1024).}
\label{tab:main_results}
\begin{tabular}{llccccc}
\toprule
\textbf{Scale} & \textbf{Method} & \textbf{Train Loss $\downarrow$} & \textbf{Val Loss $\downarrow$} & \textbf{Val PPL $\downarrow$} & \textbf{Tok/s $\uparrow$} & \textbf{Mem (GB)} \\
\midrule
\multirow{5}{*}{\shortstack{220M\\$d{=}768$\\$L{=}12$}}
 & Baseline        & 3.613 & 3.656 & 38.71 & 289k & 5.6 \\
 & AttnRes   & 3.586 & 3.622 & 37.39 & 294k & 6.2 \\
 & Full AttnRes    & 3.577 & 3.619 & 37.30 & 285k & 6.8 \\
 & \textbf{Delta Block}     & 3.577 & 3.613 & 37.08 & 306k & 5.9 \\
 & \textbf{Delta AttnRes}   & \textbf{3.563} & \textbf{3.606} & \textbf{36.83} & 246k & 7.5 \\
\midrule
\multirow{5}{*}{\shortstack{533M\\$d{=}1024$\\$L{=}24$}}
 & Baseline        & 3.428 & 3.466 & 32.00 & 243k & 13.9 \\
 & AttnRes   & 3.423 & 3.458 & 31.75 & 174k & 17.5 \\
 & Full AttnRes    & 3.422 & 3.456 & 31.68 & 103k & 24.8 \\
 & \textbf{Delta Block}     & 3.405 & 3.439 & 31.16 & 198k & 16.1 \\
 & \textbf{Delta AttnRes}   & \textbf{3.401} & \textbf{3.436} & \textbf{31.05} & 80k & 33.6 \\
\midrule
\multirow{5}{*}{\shortstack{1044M\\$d{=}1280$\\$L{=}36$}}
 & Baseline        & 3.360 & 3.391 & 29.70 & 108k & 22.5 \\
 & AttnRes   & 3.428 & 3.458 & 31.76 & 71k & 31.6 \\
 & Full AttnRes    & 3.474 & 3.508 & 33.36 & 49k & 52.0 \\
 & \textbf{Delta Block}     & \textbf{3.338} & 3.374 & 29.19 & 86k & 28.4 \\
 & \textbf{Delta AttnRes}   & 3.339 & \textbf{3.372} & \textbf{29.13} & 34k & 77.7 \\
\bottomrule
\end{tabular}
\end{table}

\paragraph{Delta methods consistently lead.} Delta AttnRes achieves the best validation PPL at all three scales: 36.83 at 220M, 31.05 at 533M, and 29.13 at 1044M. Delta Block closely follows (37.08, 31.16, 29.19), trailing by less than 0.7\% at every scale. Both delta methods beat baseline at all scales, with the gap widening as depth increases: $-$4.9\% at 220M, $-$3.0\% at 533M, and $-$1.9\% at 1044M.

\paragraph{Replacement routing degrades at scale.} AttnRes and Full AttnRes both use cumulative sources with replacement routing (and periodic reset of the residual stream). At small scale (220M), this works: AttnRes (37.39) and Full AttnRes (37.30) improve over baseline (38.71). However, at 1044M ($L{=}36$), AttnRes \emph{degrades} to 31.76 ($+$6.9\% worse than baseline's 29.70), and Full AttnRes degrades even further to 33.36 ($+$12.3\%). The degradation worsens with more frequent reset---Full AttnRes resets every layer while AttnRes resets every 4 layers---confirming that periodic reset compounds information loss at deeper scales (\S\ref{sec:analysis}).

\paragraph{Why delta routing avoids degradation.} Delta Block and Delta AttnRes differ from Block/Full AttnRes in two ways: (1)~sources are deltas ($\mathbf{v}_i = \mathbf{h}_{i+1} - \mathbf{h}_i$) rather than cumulative states, and (2)~routing is additive ($\mathbf{h} = \tilde{\mathbf{h}} + \sum \alpha_i \mathbf{v}_i$) rather than replacement ($\mathbf{h} = \sum \alpha_i \mathbf{s}_i$), which eliminates the need for reset. This realizes all three advantages of \S\ref{sec:method}: the residual stream is always preserved, every sublayer's contribution remains individually accessible, and initialization is safe. At 1044M, this converts a method that degrades ($+$6.9\%) into one that improves ($-$1.7\%).

\paragraph{Delta Block as practical default.} Per-sublayer Delta AttnRes achieves the best PPL but stores $2L$ sources, incurring $69\%$ throughput reduction and $3.5{\times}$ memory overhead at 1044M (34k tok/s, 77.7\,GB vs.\ baseline's 108k, 22.5\,GB). Delta Block amortizes this cost via block-level grouping: at 1044M it runs at 86k tok/s with 28.4\,GB ($20\%$ throughput overhead, $26\%$ memory overhead), while matching Delta AttnRes quality (29.19 vs.\ 29.13, $<$0.2\% gap). This makes Delta Block the recommended configuration for training at scale.

\subsection{Scaling From-Scratch Training}
\label{sec:scaling_8b}

The results in Table~\ref{tab:main_results} use custom model configurations. We first verify that the same trends hold on an existing architecture, then scale up to 8B parameters.

\begin{figure}[h]
\centering
\includegraphics[width=\textwidth]{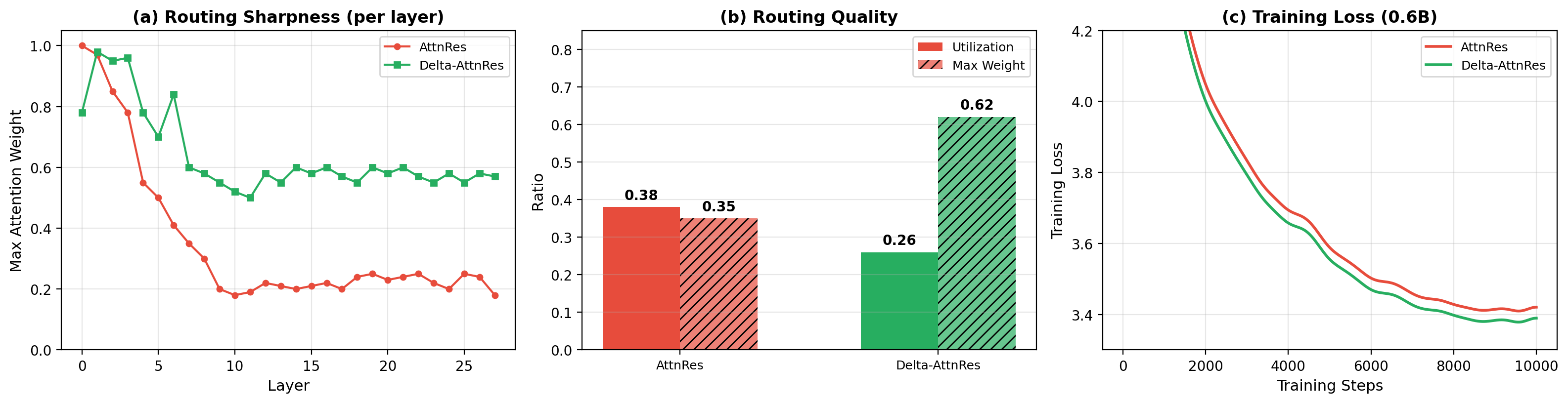}
\caption{\textbf{Routing analysis at Qwen3-0.6B scale} ($L{=}28$, $N{=}28$). (a)~Delta Block maintains sharp routing (max weight ${\sim}$0.6--1.0) while AttnRes degrades to ${\sim}$0.2 in deep layers. (b)~Delta Block achieves $1.8{\times}$ higher average max weight (0.62 vs.\ 0.35). (c)~Training loss: Delta Block (green) consistently outperforms AttnRes (red).}
\label{fig:qwen06b}
\end{figure}

\paragraph{Existing architecture: Qwen3-0.6B.} We train from scratch using the standard Qwen3-0.6B architecture ($d{=}1024$, $L{=}28$, 508M params) with per-layer routing ($N{=}L{=}28$). Delta Block improves over baseline by 2.4\% (31.45 vs.\ 32.22 PPL), while AttnRes slightly degrades (32.38), consistent with the pattern at all scales. Figure~\ref{fig:qwen06b} shows Delta Block maintains sharp routing (max weight ${\sim}$0.6) throughout depth while AttnRes collapses to ${\sim}$0.2 in deep layers.

\begin{table}[h]
\centering
\caption{From-scratch training at Qwen3-0.6B scale ($L{=}28$, $N{=}28$, 10K steps, lr $6{\times}10^{-4}$).}
\label{tab:qwen06b}
\begin{tabular}{lcccc}
\toprule
\textbf{Method} & \textbf{Val PPL $\downarrow$} & \textbf{Tok/s $\uparrow$} & \textbf{Mem (GB)} \\
\midrule
Baseline & 32.22 & 173k & 13.8 \\
AttnRes ($N{=}28$) & 32.38 & 111k & 28.4 \\
\textbf{Delta Block} ($N{=}28$) & \textbf{31.45} & 117k & 25.2 \\
\bottomrule
\end{tabular}
\end{table}

\paragraph{Scaling up: 8B parameters.} We next scale to a Qwen3-8B-sized model ($d{=}4096$, $L{=}36$, 7.57B params) trained from scratch on FineWeb-Edu for 10K steps with FSDP and gradient checkpointing on 8$\times$H100 (lr $3{\times}10^{-4}$, seq len 2048, effective batch 64).

\begin{table}[h]
\centering
\caption{\textbf{Scaling to 8B} (10K steps, FineWeb-Edu, seq len 2048, batch 64, 8$\times$H100 FSDP). Delta Block achieves the best PPL while AttnRes degrades below baseline, consistent with the 1044M trend. Tok/s and Mem measured on 8$\times$H100 (BF16, \texttt{torch.compile}, gradient checkpointing). Best in \textbf{bold}.}
\label{tab:8b}
\begin{tabular}{llccccc}
\toprule
\textbf{Scale} & \textbf{Method} & \textbf{Train Loss $\downarrow$} & \textbf{Val Loss $\downarrow$} & \textbf{Val PPL $\downarrow$} & \textbf{Tok/s $\uparrow$} & \textbf{Mem (GB)} \\
\midrule
\multirow{3}{*}{\shortstack{7.57B\\$d{=}4096$\\$L{=}36$}}
 & Baseline        & 2.918 & 2.858 & 17.43 & 21.4k & 41.6 \\
 & AttnRes         & 2.984 & 2.922 & 18.58 & 12.5k & 44.0 \\
 & \textbf{Delta Block} & \textbf{2.828} & \textbf{2.773} & \textbf{16.00} & 14.0k & 42.7 \\
\bottomrule
\end{tabular}
\end{table}

\paragraph{Delta Block leads, AttnRes degrades.} Delta Block achieves the best validation PPL (16.00), improving $-$8.2\% over baseline (17.43). AttnRes degrades to 18.58 ($+$6.6\% worse than baseline), confirming that replacement routing with cumulative sources fails at scale---the same pattern observed at 1044M ($+$6.9\%, Table~\ref{tab:main_results}).

\paragraph{Practical overhead.} Delta Block adds only 589.8K routing parameters (0.008\% of 7.57B) and incurs modest overhead: 14.0k tok/s vs.\ baseline's 21.4k ($35\%$ throughput cost) and 42.7\,GB vs.\ 41.6\,GB ($+$3\% memory). Notably, Delta Block is \emph{faster} than AttnRes (14.0k vs.\ 12.5k tok/s) and uses less memory (42.7 vs.\ 44.0\,GB), because additive routing avoids the costly hidden-state replacement and reset operations.

\subsection{Ablation: Effect of Block Size}
\label{sec:blocksize}

We ablate the number of blocks for Delta Block at 220M ($L{=}12$) and 533M ($L{=}24$). $B{=}4$ is the default used in Table~\ref{tab:main_results}.

\begin{table}[h]
\centering
\caption{Delta Block block size ablation (10K steps). Tok/s on 8$\times$H100.}
\label{tab:blocksize}
\begin{tabular}{llcccc}
\toprule
\textbf{Method} & \textbf{$B$} & \textbf{Sources} & \textbf{Val PPL $\downarrow$} & \textbf{Tok/s $\uparrow$} & \textbf{Mem (GB)} \\
\midrule
\multicolumn{6}{l}{\textit{220M ($L{=}12$)}} \\
Baseline & --- & --- & 38.71 & 289k & 5.6 \\
\textbf{Delta Block} & 1 & 2 & 37.44 & 396k & 8.7 \\
\textbf{Delta Block} & 2 & 3 & 37.08 & 318k & 8.8 \\
\textbf{Delta Block} & 4 & 5 & \textbf{36.98} & 343k & 9.1 \\
\textbf{Delta Block} & 6 & 7 & 36.92 & 315k & 9.3 \\
\textbf{Delta Block} & 12 & 13 & 37.34 & 318k & 10.2 \\
\textbf{Delta AttnRes} & --- & 25 & \textbf{36.75} & 212k & 12.4 \\
\midrule
\multicolumn{6}{l}{\textit{533M ($L{=}24$)}} \\
Baseline & --- & --- & 32.00 & 243k & 13.9 \\
\textbf{Delta Block} & 2 & 3 & 31.23 & 217k & 14.7 \\
\textbf{Delta Block} & 4 & 5 & 31.27 & 160k & 15.4 \\
\textbf{Delta Block} & 6 & 7 & \textbf{31.19} & 194k & 16.1 \\
\textbf{Delta Block} & 12 & 13 & 31.22 & 171k & 18.3 \\
\textbf{Delta Block} & 24 & 25 & 31.18 & 137k & 23.5 \\
\textbf{Delta AttnRes} & --- & 49 & \textbf{31.05} & 80k & 33.6 \\
\bottomrule
\end{tabular}
\end{table}

\paragraph{220M: sweet spot at $B{=}4$--$6$.} Delta Block improves from $B{=}1$ (37.44) to $B{=}6$ (36.92), then degrades at $B{=}12$ (37.34). Full per-sublayer Delta AttnRes (36.75) achieves the best PPL at this scale.

\paragraph{533M: robust to block size.} At 533M, Delta Block PPL is remarkably stable (31.18--31.27) across all block sizes from $B{=}2$ to $B{=}24$, suggesting that even a few delta sources capture sufficient cross-layer information at larger scale.

\paragraph{Delta Block as practical default.} Delta AttnRes achieves the best PPL at both scales but incurs significant throughput and memory overhead ($2L$ sources). Delta Block at $B{=}4$ trails by only 0.6\% at 220M while running at $1.6{\times}$ the throughput, making it the recommended configuration.

\subsection{Fine-Tuning Pretrained Models into Delta Attention Residuals}
\label{sec:finetune}

Building on existing checkpoints is common practice in modern LLM development, as pretraining costs grow prohibitively with scale. Prior work~\citep{pagliardini2024denseformer} found that adding cross-layer routing during fine-tuning fails because ``the model commits early to a loss landscape valley that does not use cross-layer weights.'' We evaluate whether delta routing's safe initialization (\S\ref{sec:method}, advantage 3) overcomes this.

\paragraph{Setup.} We fine-tune Qwen3-0.6B~\citep{qwen2025qwen3} on FineWeb-Edu~\citep{penedo2024fineweb} for 20K steps (warmup 500, cosine decay, batch size 32, 4$\times$H100). Following LoRA~\citep{hu2022lora}, we use a dual learning rate for Delta Block: $5{\times}10^{-5}$ for the pretrained transformer and $5{\times}10^{-3}$ for the AttnRes parameters, allowing the lightweight routing module to train faster while preserving pretrained knowledge. Baseline and AttnRes use a uniform lr of $5{\times}10^{-5}$. We compare: (1)~Baseline (standard fine-tuning), (2)~AttnRes~\citep{kimi2025attention}, and (3)~Delta Block + Null Source (ours). We evaluate on 8 standard benchmarks using lm-evaluation-harness~\citep{eval-harness} (0-shot): HellaSwag~\citep{zellers2019hellaswag}, ARC-Easy/Challenge~\citep{clark2018arc}, PIQA~\citep{bisk2020piqa}, WinoGrande~\citep{sakaguchi2020winogrande}, BoolQ~\citep{clark2019boolq}, MMLU~\citep{hendrycks2021mmlu}, and LAMBADA~\citep{paperno2016lambada}.

\begin{table}[h]
\centering

\caption{\textbf{Fine-tuning Qwen3-0.6B on FineWeb-Edu.} Downstream accuracy (0-shot) on 8 benchmarks. Delta Block outperforms baseline (55.6\% vs.\ 55.0\%), while AttnRes lags behind (54.1\%).}
\label{tab:finetune}
\resizebox{\textwidth}{!}{%
\begin{tabular}{lccccccccc}
\toprule
\textbf{Method} & \textbf{Hella} & \textbf{ARC-E} & \textbf{ARC-C} & \textbf{PIQA} & \textbf{Wino} & \textbf{BoolQ} & \textbf{MMLU} & \textbf{LAMB} & \textbf{Avg} \\
\midrule
Pretrained & 47.3 & 56.2 & 33.9 & 67.4 & 55.9 & 63.8 & 40.3 & 40.0 & 50.6 \\
\midrule
Baseline FT & 50.1 & 61.5 & 36.7 & 69.2 & 57.0 & 70.3 & 45.2 & 50.0 & 55.0 \\
AttnRes & 49.4 & 60.2 & 35.3 & 68.8 & 57.9 & 70.3 & 42.5 & 48.4 & 54.1 \\
\textbf{Delta Block} & 50.0 & \textbf{66.5} & \textbf{37.3} & \textbf{69.4} & 56.7 & 70.3 & 44.8 & \textbf{49.8} & \textbf{55.6} \\
\bottomrule
\end{tabular}
}
\end{table}

\paragraph{Results.} Delta Block achieves 55.6\% average accuracy, outperforming baseline (55.0\%) and AttnRes (54.1\%), with the highest ARC-Easy (66.5) and ARC-Challenge (37.3) scores.

\begin{figure}[t]
\centering
\includegraphics[width=\textwidth]{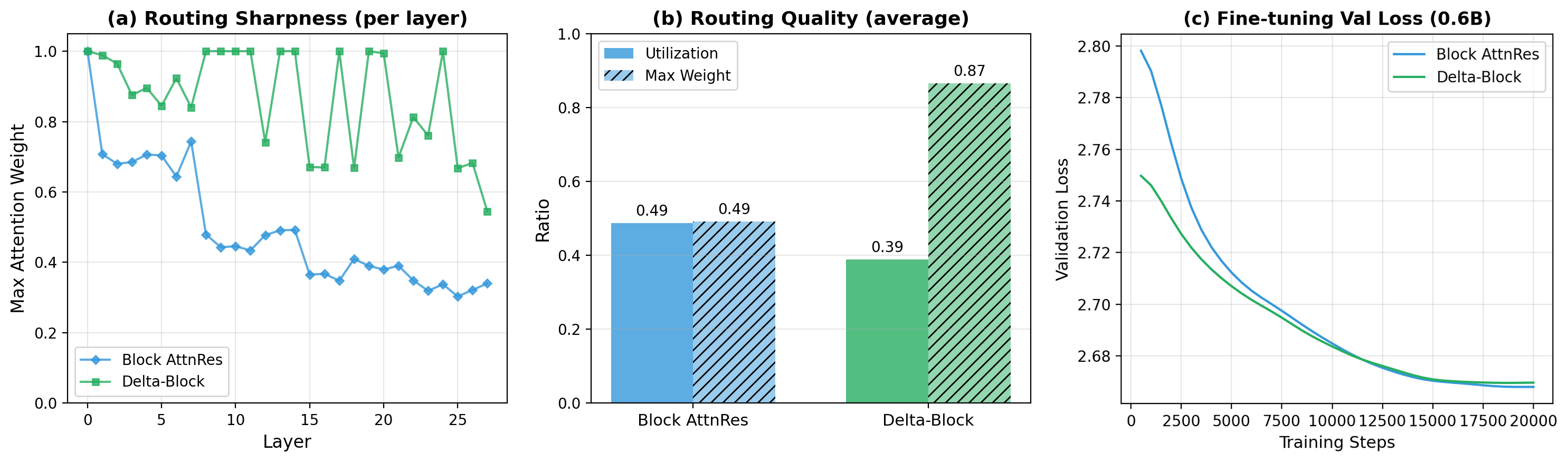}
\caption{\textbf{Routing analysis} (Qwen3-0.6B fine-tuned on FineWeb-Edu). (a)~Per-layer routing sharpness: Delta Block maintains high max attention weight (${\sim}$0.87) throughout depth, while AttnRes degrades from 0.7 to 0.3. (b)~Average routing quality: Delta Block achieves $1.8{\times}$ higher max weight (0.87 vs.\ 0.49). (c)~Validation loss: AttnRes (blue) starts higher due to initialization disruption and converges slower; Delta Block (green) outperforms baseline.}
\label{fig:finetune_curve}
\end{figure}

\paragraph{Initialization matters.} Figure~\ref{fig:finetune_curve} reveals a stark difference in training dynamics. AttnRes suffers a large loss spike at initialization (from 2.8 to 3.96) because its replacement routing ($\mathbf{h} = \sum \alpha_i \mathbf{h}_i$) returns the uniform average of all cumulative states at init---a signal fundamentally different from the pretrained residual. The model requires ${\sim}$2000 steps to recover, and the downstream gap persists. In contrast, Delta Block starts smoothly from the pretrained loss. Its additive formulation ($\mathbf{h} = \tilde{\mathbf{h}} + \sum \alpha_i \mathbf{v}_i$) preserves the residual stream by construction, and the zero-initialized null source provides an explicit identity path ($\mathbf{h} = \tilde{\mathbf{h}} + 0$), eliminating disruption entirely. This confirms that the safe initialization property (\S\ref{sec:method}, advantage 3) is essential for practical deployment atop existing checkpoints.

%% ============================================================
\section{Analysis}
\label{sec:analysis}
%% ============================================================

\subsection{Routing Collapse: Why Cumulative States Fail}
\label{sec:redundancy}

The root cause is structural: each cumulative state $\mathbf{h}_i = \mathbf{h}_0 + \sum_{j=1}^{i} \mathbf{v}_j$ is a running sum that shares most of its components with its neighbors. At Qwen3-0.6B scale ($L{=}28$) after 10K steps, AttnRes routing sharpness (max softmax weight) drops from ${\sim}$1.0 in early layers to ${\sim}$0.2 in deep layers (Figure~\ref{fig:design_space}a), meaning the model distributes attention nearly uniformly across all sources and cannot selectively access earlier representations. In contrast, Delta Block maintains sharp routing (${\sim}$0.6) throughout depth, with average max weight $1.8{\times}$ higher than AttnRes (0.62 vs.\ 0.35; Figure~\ref{fig:design_space}b). Delta sources avoid redundancy because attention and MLP outputs occupy different subspaces and capture different abstraction levels.

\subsection{Why Additive Routing Preserves Information}

This section empirically validates the three advantages identified in \S\ref{sec:method}. AttnRes \emph{replaces} the hidden state ($\mathbf{h} = \sum \alpha_i \mathbf{s}_i$) and resets at block boundaries, violating advantages 1 and 2: the residual stream is discarded, and intermediate sublayer contributions within each block are collapsed into a single sum. The information loss compounds with depth: at 1044M, AttnRes degrades to 31.76 PPL versus the 29.70 baseline ($+$6.9\%), and Full AttnRes ($N{=}L$, reset every layer) degrades even further to 33.36 ($+$12.3\%). In contrast, Delta Block (29.19) and Delta AttnRes (29.13) both improve over baseline, confirming that additive routing with delta sources avoids all three failure modes.

\subsection{Learned Routing Patterns}

\begin{figure}[t]
\centering
\includegraphics[width=\textwidth]{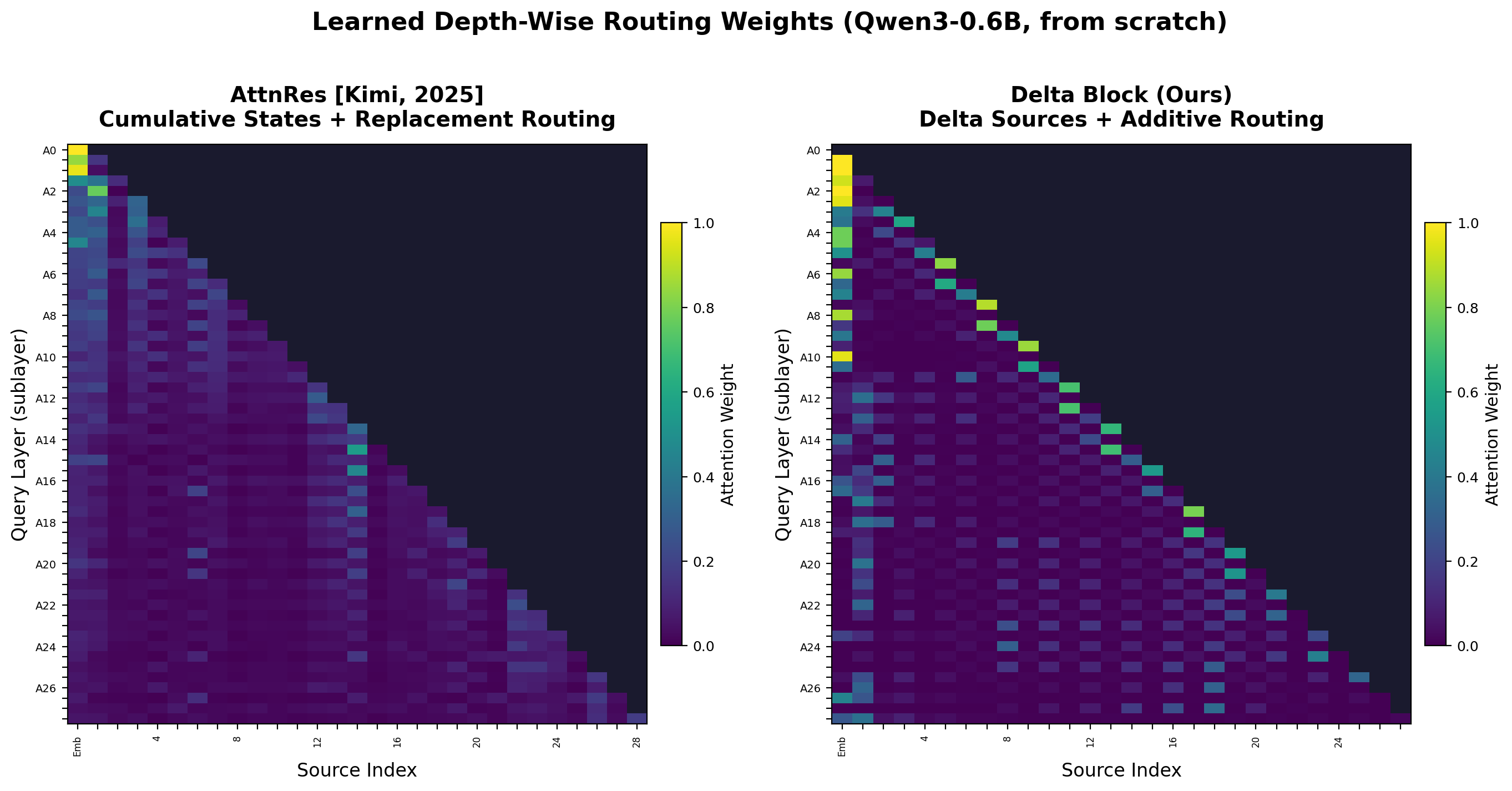}
\caption{\textbf{Learned routing weights} (Qwen3-0.6B, from scratch). \textbf{Left}: AttnRes~\citep{kimi2025attention} with cumulative states and replacement routing---attention becomes diffuse in deep layers due to source redundancy. \textbf{Right}: Delta Block (ours) with delta sources and additive routing---sharp cross-layer shortcuts, with deep layers selectively concentrating on specific early outputs.}
\label{fig:routing}
\end{figure}

Figure~\ref{fig:routing} visualizes the learned routing weights for both methods at Qwen3-0.6B scale ($L{=}28$). AttnRes with cumulative states and replacement routing (left) shows attention becoming increasingly diffuse in deeper layers, confirming the routing collapse predicted by the source redundancy analysis (\S\ref{sec:redundancy}). Delta Block with delta sources and additive routing (right) produces qualitatively different patterns:
\begin{itemize}[leftmargin=20pt]
    \item \textbf{Sharp routing}: deep layers concentrate $>$50\% weight on specific early outputs, compared to the near-uniform distribution of cumulative-state routing. This confirms that delta sources maintain discriminability throughout depth.
    \item \textbf{Embedding prominence}: the token embedding receives disproportionate attention from deep layers, consistent with progressive dilution of embedding signal under standard residuals. Additive routing allows the model to selectively re-inject this signal without disrupting the residual stream.
\end{itemize}

%% ============================================================
\section{Conclusion}
%% ============================================================

We have presented Delta Attention Residuals, which replace cumulative hidden states with per-sublayer deltas as routing sources for cross-layer connectivity. The core insight is simple: routing over \emph{what changed} ($\mathbf{v}_i = \mathbf{h}_{i+1} - \mathbf{h}_i$) rather than \emph{what accumulated} yields $3{\times}$ sharper routing (max weight ${\sim}$0.6 vs.\ ${\sim}$0.2 in deep layers). Combined with additive routing that preserves the residual stream, Delta methods achieve the best perplexity from 220M custom configurations through standard Qwen3-0.6B and Qwen3-8B architectures with the block-level variant Delta Block matching per-sublayer quality at lower overhead. At 7.6B parameters, Delta Block improves $-$8.2\% over baseline while adding only 0.008\% parameters and running faster than AttnRes. The approach is orthogonal to other architectural improvements and enables disruption-free fine-tuning of existing checkpoints.

\bibliographystyle{plainnat}
\bibliography{references}

\newpage
\appendix

\section{Related Work}
\label{app:related}

\paragraph{Cross-Layer Connections.} DenseNet~\citep{huang2017densely} concatenates all previous outputs ($O(L^2 d)$ memory). DenseFormer~\citep{pagliardini2024denseformer} uses static Depth-Weighted Average weights. Hyper-Connections~\citep{zhu2024hyperconnections} widen the residual stream to $n$ channels; mHC~\citep{deepseek2025mhc} constrains mixing to the doubly stochastic manifold. MUDDFormer~\citep{wang2024muddformer} generates dynamic, per-stream weights via MLPs, showing 2.8B matches 6.9B Pythia. Realformer~\citep{he2021realformer} passes residual attention scores across layers via additive skip connections on the attention logit matrix. Attention Residuals~\citep{kimi2025attention} use softmax attention over depth, scaling to 16B. We identify source representation as an overlooked design choice in this space.

\paragraph{Residual Stream Analysis.} \citet{veit2016residual} view residual networks as ensembles; \citet{elhage2021mathematical} formalize the residual stream. DeepNet~\citep{wang2022deepnet} rescales residual connections with a depth-dependent factor to stabilize 1000-layer transformers. Residual stream duality~\citep{zhang2026duality} shows depth attention is dual to sequence-axis sliding-window attention. Our source redundancy analysis complements these views.

\paragraph{Contrastive and Delta Methods.} A growing body of work shows that differencing redundant components yields sharper signals. Contrastive Decoding~\citep{li2023contrastive} subtracts amateur logits from expert logits to isolate scale-dependent knowledge. DoLa~\citep{chuang2024dola} contrasts early vs.\ late layers within a single model, revealing that per-layer deltas carry more distinctive factual information than cumulative representations. Proxy-Tuning~\citep{liu2024proxytuning} transfers fine-tuning residuals across model scales at decode time. Our work applies this differencing principle along the depth axis \emph{within} a single forward pass.

\paragraph{Gating and Routing.} Highway Networks~\citep{srivastava2015highway} introduced learned gating of skip connections. GLU~\citep{shazeer2020glu}, ReZero~\citep{bachlechner2020rezero}, and MoE~\citep{shazeer2017moe,fedus2022switch} route along the width dimension; Mixture of Depths~\citep{raposo2024mixture} routes tokens along depth via per-token top-$k$ decisions. We route \emph{source representations} along depth. Deep Delta Learning~\citep{zhang2026ddl} uses Householder reflections for geometric depth control; we use attention-based selection.

\section{Comparison with Original Attention Residuals}
\label{app:comparison}

We reproduce the original AttnRes pseudocode from~\citet{kimi2025attention} and analyze the key implementation differences with Delta Attention Residuals.

\begin{figure}[h]
\begin{lstlisting}[style=python]
def block_attn_res(blocks, partial_block, proj, norm):
    """Replacement routing: weighted sum replaces hidden state."""
    V = torch.stack(blocks + [partial_block])       # [N+1, B, T, D]
    K = norm(V)
    logits = einsum('d, n b t d -> n b t', proj.weight.squeeze(), K)
    h = einsum('n b t, n b t d -> b t d', logits.softmax(0), V)
    return h                                        # replaces hidden state

def forward(self, blocks, hidden_states):
    partial_block = hidden_states
    h = block_attn_res(blocks, partial_block, self.attn_res_proj, self.attn_res_norm)

    # block boundary: store and reset
    if self.layer_number % (self.block_size // 2) == 0:
        blocks.append(partial_block)
        partial_block = None                        # RESET

    attn_out = self.attn(self.attn_norm(h))
    partial_block = partial_block + attn_out if partial_block is not None else attn_out

    h = block_attn_res(blocks, partial_block, self.mlp_res_proj, self.mlp_res_norm)
    mlp_out = self.mlp(self.mlp_norm(h))
    partial_block = partial_block + mlp_out
    return blocks, partial_block
\end{lstlisting}
\caption{Original AttnRes~\citep{kimi2025attention}: replacement routing over cumulative intra-block states, with periodic reset at block boundaries.}
\label{fig:original_code}
\end{figure}

\paragraph{Three key differences.} Table~\ref{tab:diff} summarizes the design differences between the original AttnRes and our Delta variants.

\begin{table}[h]
\centering
\caption{Implementation differences between AttnRes and Delta Attention Residuals.}
\label{tab:diff}
\begin{tabular}{lccc}
\toprule
\textbf{Property} & \textbf{AttnRes} & \textbf{Delta AttnRes} & \textbf{Delta Block} \\
\midrule
Sources & cumulative $\mathbf{h}_i$ & sublayer $\mathbf{v}_i$ & block $\Delta_b$ \\
Routing & replacement & additive & additive \\
 & $h = \sum \alpha_i \mathbf{s}_i$ & $h = \tilde{\mathbf{h}} + \sum \alpha_i \mathbf{v}_i$ & $h = \tilde{\mathbf{h}} + \sum \alpha_i \Delta_b$ \\
Reset & yes (at boundaries) & no & no \\
Residual stream & lost at reset & always preserved & always preserved \\
\# Sources & $N$ blocks & $2L$ sublayers & ${\sim}L/B$ blocks \\
\bottomrule
\end{tabular}
\end{table}

\paragraph{(1) Sources: cumulative vs.\ delta.} AttnRes stores cumulative intra-block sums as sources. Since $\mathbf{h}_{i+1} = \mathbf{h}_i + \mathbf{v}_i$, adjacent cumulative sources are highly redundant, causing routing sharpness to degrade to max weight ${\sim}$0.2 in deep layers. Delta methods store the \emph{change} at each step: $\mathbf{v}_i = \mathbf{h}_{i+1} - \mathbf{h}_i$ for per-sublayer, or $\Delta_b = \mathbf{h}_{\text{current}} - \mathbf{h}_{\text{prev}}$ for block-level. These are naturally diverse, maintaining sharp routing (max weight ${\sim}$0.6).

\paragraph{(2) Routing: replacement vs.\ additive.} AttnRes \emph{replaces} the hidden state with a weighted sum of sources: $h = \sum \alpha_i \mathbf{s}_i$. This discards the current residual stream entirely. Delta methods \emph{add} to the residual stream: $h = \tilde{\mathbf{h}} + \sum \alpha_i \mathbf{v}_i$. The residual is always preserved, and at initialization (zero queries, uniform weights), the perturbation is bounded.

\paragraph{(3) Reset.} AttnRes resets \texttt{partial\_block} to zero at each block boundary, forcing the model to reconstruct the hidden state from routing alone. This loses accumulated information and makes the model fragile---especially when $N{=}L$ (each layer resets), where frequent resets compound information loss. Delta methods never reset: \texttt{hidden\_states} accumulates naturally through standard residual addition, and routing simply augments it.

\section{Delta Block: Block-Level Variant}
\label{app:block}

The \textbf{Delta Block} variant coarsens the per-sublayer sources of Delta AttnRes to block-level granularity. Rather than treating each attention output and each MLP output as a separate source ($2L$ sources for $L$ layers), Delta Block aggregates consecutive sublayer outputs within a contiguous span of $B$ layers into a single block delta
\begin{equation}
    \Delta_b = \mathbf{h}_{(b+1)B} - \mathbf{h}_{bB} = \sum_{i \in \text{block } b} \mathbf{v}_i,
    \label{eq:block_delta}
\end{equation}
yielding $B{+}1$ routing sources for a model partitioned into $B$ blocks (plus the current partial block). Routing remains additive: $\hat{\mathbf{h}}_l = \tilde{\mathbf{h}}_l + \sum_{b} \alpha_{b \to l} \cdot \Delta_b$, with $\alpha_{b \to l} = \mathrm{softmax}(\mathbf{w}_l^\top \mathrm{RMSNorm}(\Delta_b))$.

\begin{figure}[h]
\centering
\includegraphics[width=\textwidth]{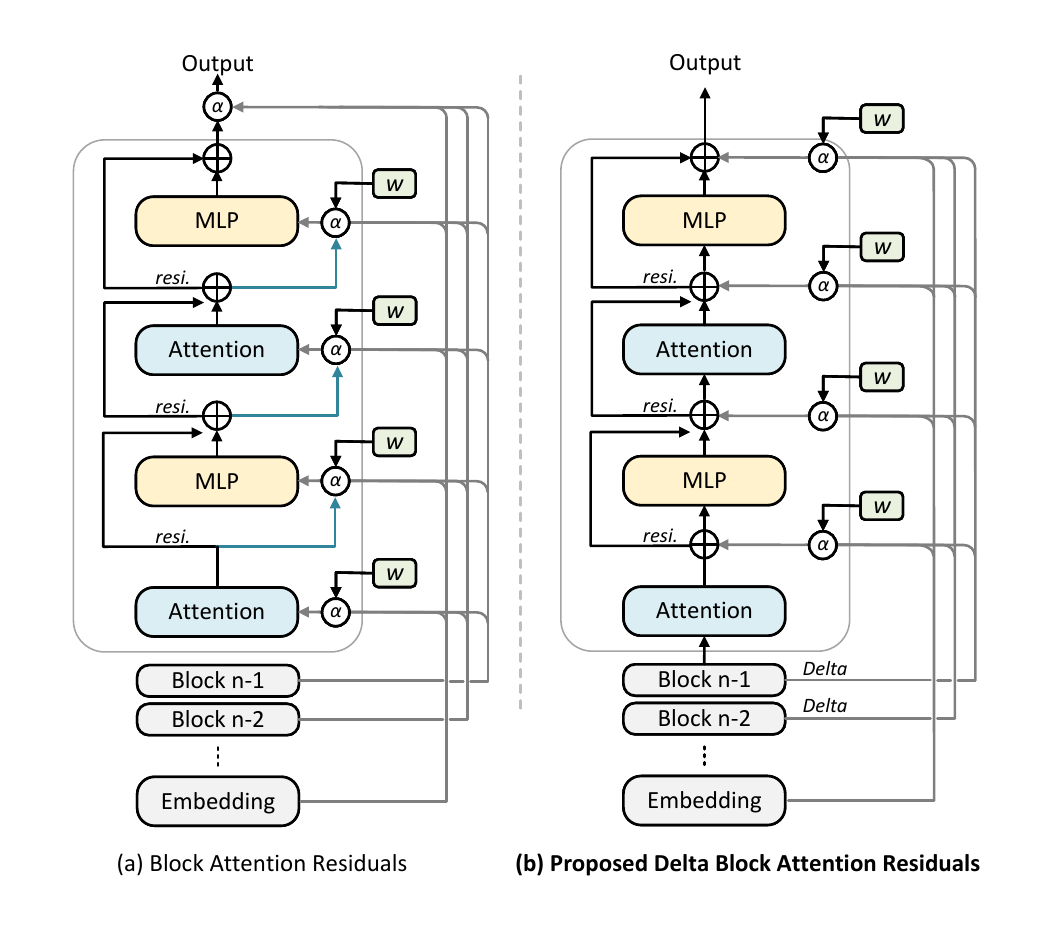}
\caption{\textbf{Delta Block schematic.} Sublayer outputs within each block of $B$ layers are summed into a single block delta $\Delta_b$, which becomes one routing source. The current (in-progress) block contributes a partial delta. Compared to per-sublayer Delta AttnRes, Delta Block reduces the number of sources from $2L$ to ${\sim}L/B$, trading routing granularity for compute and memory efficiency. The residual stream $\tilde{\mathbf{h}}_l$ is preserved throughout; routing additively augments it.}
\label{fig:delta_block}
\end{figure}

\paragraph{Properties.} Delta Block inherits the three advantages of Delta AttnRes (residual preservation, no information loss at block boundaries due to additive routing rather than replacement, and safe zero-init), and adds two block-level benefits:
\begin{itemize}[leftmargin=20pt]
    \item \textbf{Source count $\propto B$.} The number of routing sources scales with the number of blocks rather than sublayers, reducing the per-layer routing cost by a factor of ${\sim}2B$.
    \item \textbf{Coarser but still diverse.} Because each $\Delta_b$ sums structurally heterogeneous attention and MLP outputs across $B$ layers, block deltas remain mutually distinct and routing stays sharp---unlike cumulative block sources in the original AttnRes, which suffer source redundancy at depth.
\end{itemize}

\paragraph{When to prefer Delta Block.} Per-sublayer Delta AttnRes offers maximum routing expressivity and is preferred at small-to-medium scales (\S\ref{sec:experiments}). Delta Block trades a small amount of routing granularity for lower overhead and is preferred when source-count overhead dominates, e.g., at very deep models or under tight memory budgets. Empirically, Delta Block matches Delta AttnRes on most evaluations while using fewer routing sources.

\newpage
\section*{NeurIPS Paper Checklist}

\begin{enumerate}

\item {\bf Claims}
    \item[] Question: Do the main claims made in the abstract and introduction accurately reflect the paper's contributions and scope?
    \item[] Answer: \answerYes{}
    \item[] Justification: The abstract and introduction claim three contributions (routing collapse identification, Delta Attention Residuals, fine-tuning capability), all supported by experiments in \S\ref{sec:experiments}--\ref{sec:finetune}.
    \item[] Guidelines:
    \begin{itemize}
        \item The answer \answerNA{} means that the abstract and introduction do not include the claims made in the paper.
        \item The abstract and/or introduction should clearly state the claims made, including the contributions made in the paper and important assumptions and limitations. A \answerNo{} or \answerNA{} answer to this question will not be perceived well by the reviewers.
        \item The claims made should match theoretical and experimental results, and reflect how much the results can be expected to generalize to other settings.
        \item It is fine to include aspirational goals as motivation as long as it is clear that these goals are not attained by the paper.
    \end{itemize}

\item {\bf Limitations}
    \item[] Question: Does the paper discuss the limitations of the work performed by the authors?
    \item[] Answer: \answerYes{}
    \item[] Justification: Controlled comparisons are conducted at scales up to 7.6B parameters (\S\ref{sec:scaling_8b}), but the 8B experiment evaluates only Baseline, AttnRes, and Delta Block (not Full AttnRes or Delta AttnRes due to memory constraints). Per-sublayer Delta AttnRes incurs significant memory overhead at large scale (\S\ref{sec:experiments}). Fine-tuning is evaluated on a single base model (Qwen3-0.6B).
    \item[] Guidelines:
    \begin{itemize}
        \item The answer \answerNA{} means that the paper has no limitation while the answer \answerNo{} means that the paper has limitations, but those are not discussed in the paper.
        \item The authors are encouraged to create a separate ``Limitations'' section in their paper.
        \item The paper should point out any strong assumptions and how robust the results are to violations of these assumptions (e.g., independence assumptions, noiseless settings, model well-specification, asymptotic approximations only holding locally). The authors should reflect on how these assumptions might be violated in practice and what the implications would be.
        \item The authors should reflect on the scope of the claims made, e.g., if the approach was only tested on a few datasets or with a few runs. In general, empirical results often depend on implicit assumptions, which should be articulated.
        \item The authors should reflect on the factors that influence the performance of the approach. For example, a facial recognition algorithm may perform poorly when image resolution is low or images are taken in low lighting. Or a speech-to-text system might not be used reliably to provide closed captions for online lectures because it fails to handle technical jargon.
        \item The authors should discuss the computational efficiency of the proposed algorithms and how they scale with dataset size.
        \item If applicable, the authors should discuss possible limitations of their approach to address problems of privacy and fairness.
        \item While the authors might fear that complete honesty about limitations might be used by reviewers as grounds for rejection, a worse outcome might be that reviewers discover limitations that aren't acknowledged in the paper. The authors should use their best judgment and recognize that individual actions in favor of transparency play an important role in developing norms that preserve the integrity of the community. Reviewers will be specifically instructed to not penalize honesty concerning limitations.
    \end{itemize}

\item {\bf Theory assumptions and proofs}
    \item[] Question: For each theoretical result, does the paper provide the full set of assumptions and a complete (and correct) proof?
    \item[] Answer: \answerYes{}
    \item[] Justification: Proposition 1 (\S\ref{sec:what}) states its assumptions explicitly (running-sum structure of cumulative states) and provides both a formal argument and empirical validation.
    \item[] Guidelines:
    \begin{itemize}
        \item The answer \answerNA{} means that the paper does not include theoretical results.
        \item All the theorems, formulas, and proofs in the paper should be numbered and cross-referenced.
        \item All assumptions should be clearly stated or referenced in the statement of any theorems.
        \item The proofs can either appear in the main paper or the supplemental material, but if they appear in the supplemental material, the authors are encouraged to provide a short proof sketch to provide intuition.
        \item Inversely, any informal proof provided in the core of the paper should be complemented by formal proofs provided in appendix or supplemental material.
        \item Theorems and Lemmas that the proof relies upon should be properly referenced.
    \end{itemize}

    \item {\bf Experimental result reproducibility}
    \item[] Question: Does the paper fully disclose all the information needed to reproduce the main experimental results of the paper to the extent that it affects the main claims and/or conclusions of the paper (regardless of whether the code and data are provided or not)?
    \item[] Answer: \answerYes{}
    \item[] Justification: \S\ref{sec:experiments} specifies all hyperparameters, optimizer settings, hardware, and training budget. Complete pseudocode is provided in Figure~\ref{fig:code}. Code is available at \url{https://github.com/wdlctc/delta-attention-residuals-code}.
    \item[] Guidelines:
    \begin{itemize}
        \item The answer \answerNA{} means that the paper does not include experiments.
        \item If the paper includes experiments, a \answerNo{} answer to this question will not be perceived well by the reviewers: Making the paper reproducible is important, regardless of whether the code and data are provided or not.
        \item If the contribution is a dataset and\slash or model, the authors should describe the steps taken to make their results reproducible or verifiable.
        \item Depending on the contribution, reproducibility can be accomplished in various ways. For example, if the contribution is a novel architecture, describing the architecture fully might suffice, or if the contribution is a specific model and empirical evaluation, it may be necessary to either make it possible for others to replicate the model with the same dataset, or provide access to the model. In general. releasing code and data is often one good way to accomplish this, but reproducibility can also be provided via detailed instructions for how to replicate the results, access to a hosted model (e.g., in the case of a large language model), releasing of a model checkpoint, or other means that are appropriate to the research performed.
        \item While NeurIPS does not require releasing code, the conference does require all submissions to provide some reasonable avenue for reproducibility, which may depend on the nature of the contribution. For example
        \begin{enumerate}
            \item If the contribution is primarily a new algorithm, the paper should make it clear how to reproduce that algorithm.
            \item If the contribution is primarily a new model architecture, the paper should describe the architecture clearly and fully.
            \item If the contribution is a new model (e.g., a large language model), then there should either be a way to access this model for reproducing the results or a way to reproduce the model (e.g., with an open-source dataset or instructions for how to construct the dataset).
            \item We recognize that reproducibility may be tricky in some cases, in which case authors are welcome to describe the particular way they provide for reproducibility. In the case of closed-source models, it may be that access to the model is limited in some way (e.g., to registered users), but it should be possible for other researchers to have some path to reproducing or verifying the results.
        \end{enumerate}
    \end{itemize}

\item {\bf Open access to data and code}
    \item[] Question: Does the paper provide open access to the data and code, with sufficient instructions to faithfully reproduce the main experimental results, as described in supplemental material?
    \item[] Answer: \answerYes{}
    \item[] Justification: We use the publicly available FineWeb-Edu dataset and Qwen3 model family. Code is available at \url{https://github.com/wdlctc/delta-attention-residuals-code}.
    \item[] Guidelines:
    \begin{itemize}
        \item The answer \answerNA{} means that paper does not include experiments requiring code.
        \item Please see the NeurIPS code and data submission guidelines (\url{https://neurips.cc/public/guides/CodeSubmissionPolicy}) for more details.
        \item While we encourage the release of code and data, we understand that this might not be possible, so \answerNo{} is an acceptable answer. Papers cannot be rejected simply for not including code, unless this is central to the contribution (e.g., for a new open-source benchmark).
        \item The instructions should contain the exact command and environment needed to run to reproduce the results. See the NeurIPS code and data submission guidelines (\url{https://neurips.cc/public/guides/CodeSubmissionPolicy}) for more details.
        \item The authors should provide instructions on data access and preparation, including how to access the raw data, preprocessed data, intermediate data, and generated data, etc.
        \item The authors should provide scripts to reproduce all experimental results for the new proposed method and baselines. If only a subset of experiments are reproducible, they should state which ones are omitted from the script and why.
        \item At submission time, to preserve anonymity, the authors should release anonymized versions (if applicable).
        \item Providing as much information as possible in supplemental material (appended to the paper) is recommended, but including URLs to data and code is permitted.
    \end{itemize}

\item {\bf Experimental setting/details}
    \item[] Question: Does the paper specify all the training and test details (e.g., data splits, hyperparameters, how they were chosen, type of optimizer) necessary to understand the results?
    \item[] Answer: \answerYes{}
    \item[] Justification: \S\ref{sec:experiments} provides optimizer (AdamW), learning rate schedule (cosine with warmup), batch size, sequence length, hardware (8$\times$H100), and training steps. Fine-tuning details are in \S\ref{sec:finetune}. Downstream evaluation uses lm-evaluation-harness with 0-shot.
    \item[] Guidelines:
    \begin{itemize}
        \item The answer \answerNA{} means that the paper does not include experiments.
        \item The experimental setting should be presented in the core of the paper to a level of detail that is necessary to appreciate the results and make sense of them.
        \item The full details can be provided either with the code, in appendix, or as supplemental material.
    \end{itemize}

\item {\bf Experiment statistical significance}
    \item[] Question: Does the paper report error bars suitably and correctly defined or other appropriate information about the statistical significance of the experiments?
    \item[] Answer: \answerNo{}
    \item[] Justification: Due to computational cost (each run requires 8$\times$H100 GPUs), we report single-run results following standard practice in language model scaling studies. We mitigate this by evaluating across five scales (220M, 533M, 1044M, Qwen3-0.6B, Qwen3-8B) and showing consistent trends.
    \item[] Guidelines:
    \begin{itemize}
        \item The answer \answerNA{} means that the paper does not include experiments.
        \item The authors should answer \answerYes{} if the results are accompanied by error bars, confidence intervals, or statistical significance tests, at least for the experiments that support the main claims of the paper.
        \item The factors of variability that the error bars are capturing should be clearly stated (for example, train/test split, initialization, random drawing of some parameter, or overall run with given experimental conditions).
        \item The method for calculating the error bars should be explained (closed form formula, call to a library function, bootstrap, etc.)
        \item The assumptions made should be given (e.g., Normally distributed errors).
        \item It should be clear whether the error bar is the standard deviation or the standard error of the mean.
        \item It is OK to report 1-sigma error bars, but one should state it. The authors should preferably report a 2-sigma error bar than state that they have a 96\% CI, if the hypothesis of Normality of errors is not verified.
        \item For asymmetric distributions, the authors should be careful not to show in tables or figures symmetric error bars that would yield results that are out of range (e.g., negative error rates).
        \item If error bars are reported in tables or plots, the authors should explain in the text how they were calculated and reference the corresponding figures or tables in the text.
    \end{itemize}

\item {\bf Experiments compute resources}
    \item[] Question: For each experiment, does the paper provide sufficient information on the computer resources (type of compute workers, memory, time of execution) needed to reproduce the experiments?
    \item[] Answer: \answerYes{}
    \item[] Justification: \S\ref{sec:experiments} reports hardware (8$\times$H100 80GB), precision (BF16), throughput (Tok/s), and peak GPU memory for all configurations in Table~\ref{tab:main_results}. Fine-tuning uses 4$\times$H100.
    \item[] Guidelines:
    \begin{itemize}
        \item The answer \answerNA{} means that the paper does not include experiments.
        \item The paper should indicate the type of compute workers CPU or GPU, internal cluster, or cloud provider, including relevant memory and storage.
        \item The paper should provide the amount of compute required for each of the individual experimental runs as well as estimate the total compute.
        \item The paper should disclose whether the full research project required more compute than the experiments reported in the paper (e.g., preliminary or failed experiments that didn't make it into the paper).
    \end{itemize}

\item {\bf Code of ethics}
    \item[] Question: Does the research conducted in the paper conform, in every respect, with the NeurIPS Code of Ethics \url{https://neurips.cc/public/EthicsGuidelines}?
    \item[] Answer: \answerYes{}
    \item[] Justification: This work studies architectural modifications to transformers and does not raise ethical concerns beyond those common to language model research.
    \item[] Guidelines:
    \begin{itemize}
        \item The answer \answerNA{} means that the authors have not reviewed the NeurIPS Code of Ethics.
        \item If the authors answer \answerNo, they should explain the special circumstances that require a deviation from the Code of Ethics.
        \item The authors should make sure to preserve anonymity (e.g., if there is a special consideration due to laws or regulations in their jurisdiction).
    \end{itemize}

\item {\bf Broader impacts}
    \item[] Question: Does the paper discuss both potential positive societal impacts and negative societal impacts of the work performed?
    \item[] Answer: \answerNA{}
    \item[] Justification: This is foundational architecture research on residual connection design. It does not introduce new capabilities or applications beyond what existing transformers already provide, and there is no direct path to specific negative applications.
    \item[] Guidelines:
    \begin{itemize}
        \item The answer \answerNA{} means that there is no societal impact of the work performed.
        \item If the authors answer \answerNA{} or \answerNo, they should explain why their work has no societal impact or why the paper does not address societal impact.
        \item Examples of negative societal impacts include potential malicious or unintended uses (e.g., disinformation, generating fake profiles, surveillance), fairness considerations (e.g., deployment of technologies that could make decisions that unfairly impact specific groups), privacy considerations, and security considerations.
        \item The conference expects that many papers will be foundational research and not tied to particular applications, let alone deployments. However, if there is a direct path to any negative applications, the authors should point it out. For example, it is legitimate to point out that an improvement in the quality of generative models could be used to generate Deepfakes for disinformation. On the other hand, it is not needed to point out that a generic algorithm for optimizing neural networks could enable people to train models that generate Deepfakes faster.
        \item The authors should consider possible harms that could arise when the technology is being used as intended and functioning correctly, harms that could arise when the technology is being used as intended but gives incorrect results, and harms following from (intentional or unintentional) misuse of the technology.
        \item If there are negative societal impacts, the authors could also discuss possible mitigation strategies (e.g., gated release of models, providing defenses in addition to attacks, mechanisms for monitoring misuse, mechanisms to monitor how a system learns from feedback over time, improving the efficiency and accessibility of ML).
    \end{itemize}

\item {\bf Safeguards}
    \item[] Question: Does the paper describe safeguards that have been put in place for responsible release of data or models that have a high risk for misuse (e.g., pre-trained language models, image generators, or scraped datasets)?
    \item[] Answer: \answerNA{}
    \item[] Justification: The models trained in this work are small-scale (up to 1B parameters) and trained on publicly available data. They do not pose risks beyond those of existing publicly available language models.
    \item[] Guidelines:
    \begin{itemize}
        \item The answer \answerNA{} means that the paper poses no such risks.
        \item Released models that have a high risk for misuse or dual-use should be released with necessary safeguards to allow for controlled use of the model, for example by requiring that users adhere to usage guidelines or restrictions to access the model or implementing safety filters.
        \item Datasets that have been scraped from the Internet could pose safety risks. The authors should describe how they avoided releasing unsafe images.
        \item We recognize that providing effective safeguards is challenging, and many papers do not require this, but we encourage authors to take this into account and make a best faith effort.
    \end{itemize}

\item {\bf Licenses for existing assets}
    \item[] Question: Are the creators or original owners of assets (e.g., code, data, models), used in the paper, properly credited and are the license and terms of use explicitly mentioned and properly respected?
    \item[] Answer: \answerYes{}
    \item[] Justification: We cite FineWeb-Edu, Qwen3, lm-evaluation-harness, and all benchmark datasets used. FineWeb-Edu is released under Apache 2.0; Qwen3 models are publicly available.
    \item[] Guidelines:
    \begin{itemize}
        \item The answer \answerNA{} means that the paper does not use existing assets.
        \item The authors should cite the original paper that produced the code package or dataset.
        \item The authors should state which version of the asset is used and, if possible, include a URL.
        \item The name of the license (e.g., CC-BY 4.0) should be included for each asset.
        \item For scraped data from a particular source (e.g., website), the copyright and terms of service of that source should be provided.
        \item If assets are released, the license, copyright information, and terms of use in the package should be provided. For popular datasets, \url{paperswithcode.com/datasets} has curated licenses for some datasets. Their licensing guide can help determine the license of a dataset.
        \item For existing datasets that are re-packaged, both the original license and the license of the derived asset (if it has changed) should be provided.
        \item If this information is not available online, the authors are encouraged to reach out to the asset's creators.
    \end{itemize}

\item {\bf New assets}
    \item[] Question: Are new assets introduced in the paper well documented and is the documentation provided alongside the assets?
    \item[] Answer: \answerYes{}
    \item[] Justification: We provide complete pseudocode (Figure~\ref{fig:code}) and will release code and model checkpoints under an open-source license upon publication.
    \item[] Guidelines:
    \begin{itemize}
        \item The answer \answerNA{} means that the paper does not release new assets.
        \item Researchers should communicate the details of the dataset\slash code\slash model as part of their submissions via structured templates. This includes details about training, license, limitations, etc.
        \item The paper should discuss whether and how consent was obtained from people whose asset is used.
        \item At submission time, remember to anonymize your assets (if applicable). You can either create an anonymized URL or include an anonymized zip file.
    \end{itemize}

\item {\bf Crowdsourcing and research with human subjects}
    \item[] Question: For crowdsourcing experiments and research with human subjects, does the paper include the full text of instructions given to participants and screenshots, if applicable, as well as details about compensation (if any)?
    \item[] Answer: \answerNA{}
    \item[] Justification: This work does not involve crowdsourcing or human subjects.
    \item[] Guidelines:
    \begin{itemize}
        \item The answer \answerNA{} means that the paper does not involve crowdsourcing nor research with human subjects.
        \item Including this information in the supplemental material is fine, but if the main contribution of the paper involves human subjects, then as much detail as possible should be included in the main paper.
        \item According to the NeurIPS Code of Ethics, workers involved in data collection, curation, or other labor should be paid at least the minimum wage in the country of the data collector.
    \end{itemize}

\item {\bf Institutional review board (IRB) approvals or equivalent for research with human subjects}
    \item[] Question: Does the paper describe potential risks incurred by study participants, whether such risks were disclosed to the subjects, and whether Institutional Review Board (IRB) approvals (or an equivalent approval/review based on the requirements of your country or institution) were obtained?
    \item[] Answer: \answerNA{}
    \item[] Justification: This work does not involve human subjects research.
    \item[] Guidelines:
    \begin{itemize}
        \item The answer \answerNA{} means that the paper does not involve crowdsourcing nor research with human subjects.
        \item Depending on the country in which research is conducted, IRB approval (or equivalent) may be required for any human subjects research. If you obtained IRB approval, you should clearly state this in the paper.
        \item We recognize that the procedures for this may vary significantly between institutions and locations, and we expect authors to adhere to the NeurIPS Code of Ethics and the guidelines for their institution.
        \item For initial submissions, do not include any information that would break anonymity (if applicable), such as the institution conducting the review.
    \end{itemize}

\item {\bf Declaration of LLM usage}
    \item[] Question: Does the paper describe the usage of LLMs if it is an important, original, or non-standard component of the core methods in this research? Note that if the LLM is used only for writing, editing, or formatting purposes and does \emph{not} impact the core methodology, scientific rigor, or originality of the research, declaration is not required.
    \item[] Answer: \answerNA{}
    \item[] Justification: LLMs were not used as a component of the core methodology. The research focuses on architectural modifications to transformer residual connections.
    \item[] Guidelines:
    \begin{itemize}
        \item The answer \answerNA{} means that the core method development in this research does not involve LLMs as any important, original, or non-standard components.
        \item Please refer to our LLM policy in the NeurIPS handbook for what should or should not be described.
    \end{itemize}

\end{enumerate}

\end{document}